\pdfoutput=1

\documentclass[11pt]{article}

\usepackage[preprint]{acl}

\usepackage{times}
\usepackage{latexsym}

\usepackage[T1]{fontenc}

\usepackage[utf8]{inputenc}

\usepackage{microtype}

\usepackage{inconsolata}

\usepackage{graphicx}
\usepackage{bm}
\usepackage{amsmath}
\usepackage{algorithm}
\usepackage{algorithmic}
\usepackage{subcaption}
\usepackage{amssymb}
\usepackage{ulem}

\title{QuaDMix: Quality-Diversity Balanced Data Selection for Efficient LLM Pretraining}

\author{
 \textbf{Fengze Liu\textsuperscript{1,*}},
 \textbf{Weidong Zhou\textsuperscript{1}},
 \textbf{Binbin Liu\textsuperscript{1}},
 \textbf{Zhimiao Yu\textsuperscript{1}},
\\
 \textbf{Yifan Zhang\textsuperscript{1}},
 \textbf{Haobin Lin\textsuperscript{1}},
 \textbf{Yifeng Yu\textsuperscript{1}},
  \textbf{Bingni Zhang\textsuperscript{1}},
\\
\textbf{Xiaohuan Zhou \textsuperscript{1,*}},
 \textbf{Taifeng Wang\textsuperscript{1}},
 \textbf{Yong Cao\textsuperscript{1}},
\\
\\
 \textsuperscript{1}ByteDance
\\
 \small{
 \{fengze.liu, zhouweidong.66, liubinbin.22, yuzhimiao, zzhangyifan, linhaobin.theseeker,} \\
 \small{
 yuyifeng.oscar, bingni.zhang, zhouxiaohuan, wangtaifeng, yongc\}@bytedance.com
  }
}

\begin{document}
\maketitle
\begin{abstract}

Quality and diversity are two critical metrics for the training data of large language models (LLMs), positively impacting performance. Existing studies often optimize these metrics separately, typically by first applying quality filtering and then adjusting data proportions. However, these approaches overlook the inherent trade-off between quality and diversity, necessitating their joint consideration. Given a fixed training quota, it’s essential to evaluate both the quality of each data point and its complementary effect on the overall dataset. In this paper, we introduce a unified data selection framework called QuaDMix, which automatically optimizes the data distribution for LLM pretraining while balancing both quality and diversity. Specifically, we first propose multiple criteria to measure data quality and employ domain classification to distinguish data points, thereby measuring overall diversity. QuaDMix then employs a unified parameterized data sampling function that determines the sampling probability of each data point based on these quality and diversity related labels. To accelerate the search for the optimal parameters involved in the QuaDMix framework, we conduct simulated experiments on smaller models and use LightGBM for parameters searching, inspired by the RegMix method. Our experiments across diverse models and datasets demonstrate that QuaDMix achieves an average performance improvement of 7.2\% across multiple benchmarks. These results outperform the independent strategies for quality and diversity, highlighting the necessity and the framework’s ability to balance data quality and diversity.

\end{abstract}

\section{Introduction}

The efficiency and preference of pretraining large language models are significantly influenced by the characteristics of the training corpus \citep{Brown:2023,Chowdhery:2023,Longpre:2024}. There is evidence from existing research suggesting that the model performance can be improved through the curation of high-quality data \citep{QuRating,DSIR,askllm}, the application of data deduplication and diversification strategies \citep{SemDeDup,D4}, and the careful balancing of data distribution across various domains and topics \citep{regmix,doremi}. Nevertheless, identifying optimal configuration of combining those factors remains an open challenge, due to complex interplay between data quality and diversity, which has yet to be fully understood.

\begin{figure}[t] 
    \centering
    \includegraphics[width=0.45\textwidth]{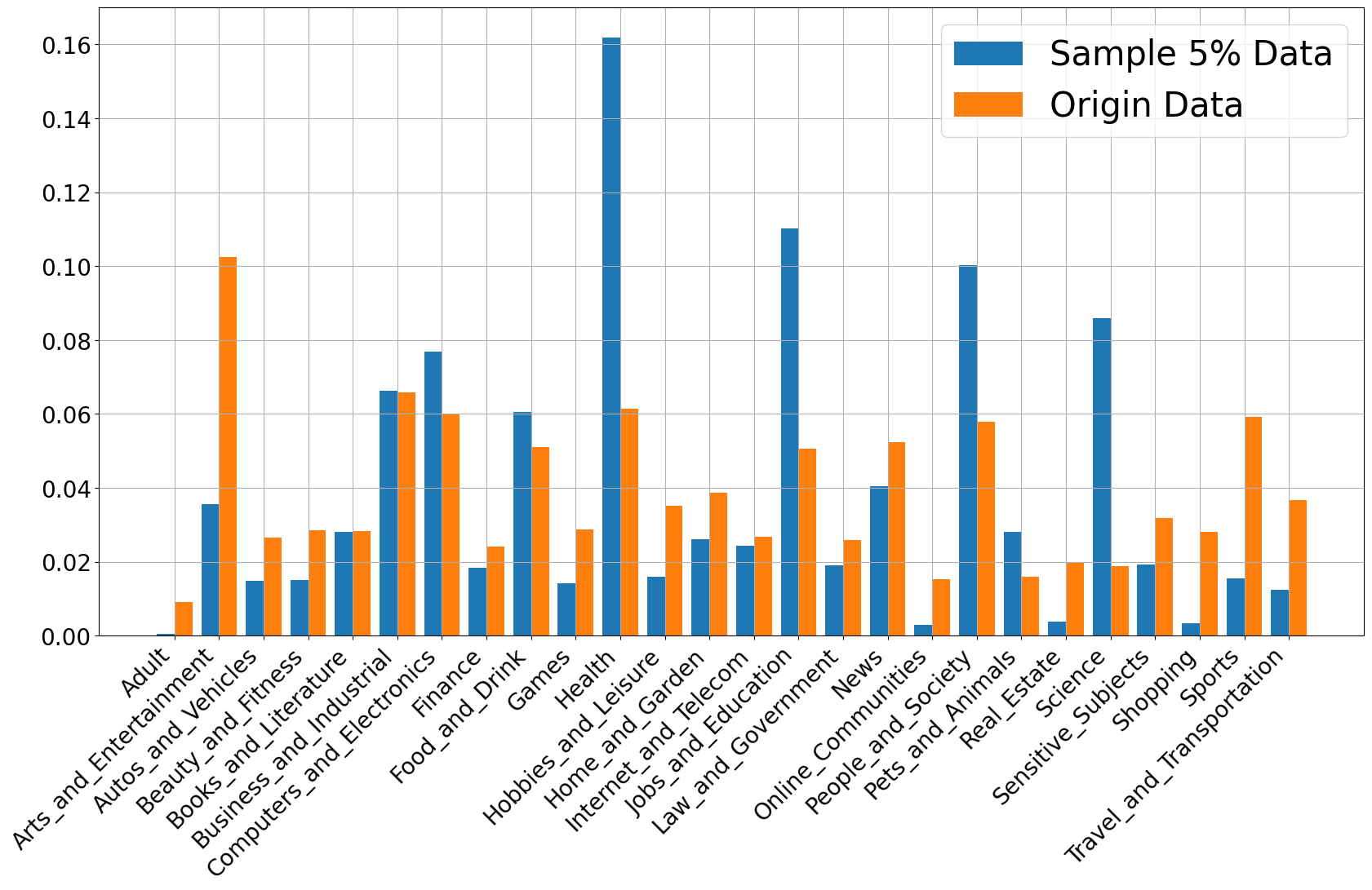} 
    \caption{The distribution change of data selected with Fineweb-edu Classifier. With the top5\% documents selected, the ratio of certain domains including Health, Jobs and Education, increases for a large margin compared with original data }
    \label{fig:fwb_distribution} 
\end{figure}

There remains two major challenges to identify the optimal data selection strategy. Firstly, the definition of quality and diversity is ambiguous. Previous research has proposed various definitions of quality criteria, including factors such as regular expression \citep{refinedweb,ccnet}, educational value \citep{fineweb-edu}, similarity to instruction tuning data \citep{dclm}, etc, each emphasizing only a specific aspect of the data. On the other hand, approaches like \citep{regmix,SemDeDup} optimize the data mixtures for more effective training, indicating that a better diversity is not necessarily uniform distribution.
Secondly, there exists interplay between data quality and diversity. The choice of quality criteria affects the distribution of selected data as illustrated in Figure \ref{fig:fwb_distribution}, due to inherent biases in different criteria. Meanwhile, changing of data mixtures influences the data quality, as the quality level differs across different domains. Also, since the high quality data is limited, the trade-off between better quality or diversity is inevitable, which is not feasible by optimizing only for data quality or diversity. How to jointly optimize the data distribution together with the selection of quality criteria remains another unsolved issue.

To address these challenges, we propose a unified data selection framework, QuaDMix, which simultaneously manages data quality and diversity. Firstly, we apply several quality scorers and domain classification on each document in the training corpus, to measure the data quality and diversity. Then a parameterized function is designed to determine the sampling frequency for each document based on those quality and domain labels. Specifically, an aggregated quality score is first computed by weighted averaging the quality scores, where the weights are controlled by adjustable parameters. Then a parameterized sampling function takes the aggregated quality score as input and calculate the sampling frequency, where data with higher quality is assigned with more frequency and the parameters affect how the frequency decreases as the quality diminishes. Here we take the assumption that training samples with higher quality worth sampled for more times. We assign independent parameters for data across different domains to control the diversity via parameters. To find the optimal parameters among the numerous parameter space, we employ a two-step approach inspired by \citep{regmix}. First, we train a set of small models on datasets sampled using QuaDMix with various parameter configurations, as an approximation for the performance of larger models. Next, we train a regression model to fit the performance results from this limited set of small models. This regression model is then used to predict the performance for unseen parameter configurations, providing an efficient way to explore the parameter space without exhaustive large-scale training.


To validate the effectiveness of QuaDMix, we train 3000 models with 1M parameters for 1B tokens, each using data sampled from RefinedWeb \citep{refinedweb} with various QuaDMix parameters. The optimal parameter configuration is then determined by searching the input space of a trained LightGBM regressor\citep{lightgbm}. We then evaluate different pretraining data selection methods on models with 530M parameters. The optimal configuration identified by QuaDMix achieves superior performance on an aggregated benchmark. Our results also reveal the following insights: (1) Different quality criteria exhibit trade-offs across downstream tasks, but appropriately merging these criteria yields consistent improvements across tasks by leveraging complementary information. (2) The optimal data mixture varies under different quality criteria, indicating the importance of jointly optimizing both the quality and diversity. (3) The target of regression model can guide the preference for specific downstream tasks, enabling task-focused data selection.
\begin{figure*}[t] 
    \centering
    \includegraphics[width=0.95\textwidth]{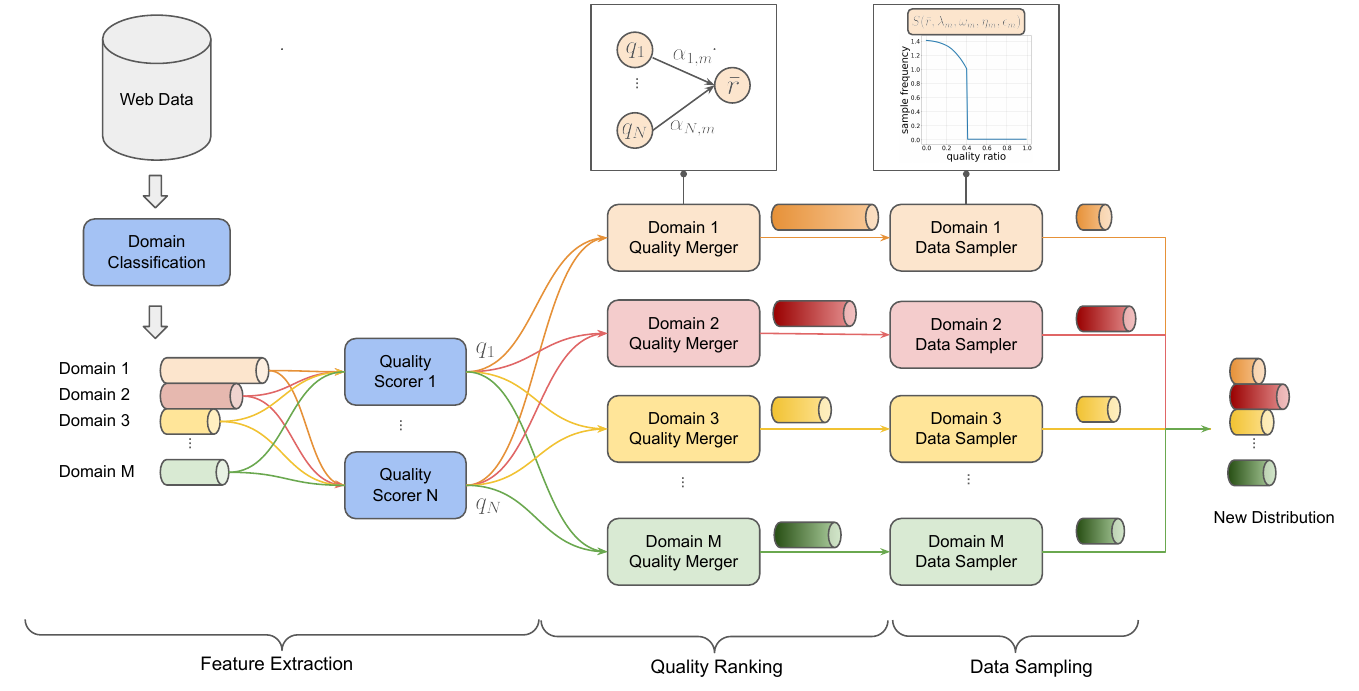} 
    \caption{The overall design of QuaDMix. First we extract the data features using classifier and quality scores (QS). Then we calculate quality rank for each domain with the merging parameters. Finally the sampling functions controlled by sampling parameters are applied to generate the final output data.}
    \label{fig:image_label} 
\end{figure*}
\section{Related Work}
\subsection{Pretraining Data Selection}

Data quality, diversity, and coverage are critical factors for ensuring the efficiency and generalizability of large language models \citep{InstructionPT, llama, Chowdhery:2023}.

To improve data quality, rule-based filtering techniques are commonly employed \citep{ROOT, RedPajama, refinedweb, C4}. These methods use handcrafted heuristics, such as removing terminal marks, detecting sentence repetitions, and enforcing length constraints, to exclude low-quality data. While these rules effectively filter out noisy data from the training corpus, they fail to capture semantic-level information, which is crucial for more refined data selection. Alternative approaches aim to address this limitation. For instance, \citep{ccnet, LIMO, Tristan:2024} use model perplexity as a measure of data quality, while \citep{Rho} apply token-level selection by re-weighting the loss across tokens. \citep{DSIR} utilize n-gram features to quantify data importance and sample accordingly. Discriminator-based methods \citep{Brown:2023, glam, pilecc, Dolma, dclm} select data by comparing it to predefined high-quality datasets, such as Wikipedia or instruction-tuning datasets. However, how much these predefined datasets represent for high-quality relies on empirical judgement. More recently, approaches like \citep{phi1, askllm, QuRating, fineweb-edu} leverage large language models (e.g., GPT-4) to evaluate and filter data based on designed prompts that emphasize various dimensions of value, offering a more nuanced way to define and curate high-quality data.

To optimize data distribution, various methods leverage clustering and representativeness to achieve deduplication and diversification. For example, \citep{SemDeDup, Clusterclip, D4} employ data clustering techniques to identify and select representative data points, ensuring both diversity and efficiency in the training corpus. Other approaches estimate optimal data mixtures through iterative modeling. \citep{doremi} first train a small reference model and subsequently optimize the worst-case loss across domains by training a proxy model to identify the optimal data mixture. Similarly, \citep{Multi-Agent, MATES, DoGE, OptimalControl} calculate influence scores by tracking first-order gradients on an evaluation set, thereby identifying the most valuable data for training. Additionally, \citep{regmix, DML} simulate the performance of different data mixtures by training a series of proxy models, enabling the prediction of large-model performance with low compute cost.

\subsection{Scaling Laws}
Neural Scaling Laws have been shown to effectively predict performance across varying training budgets, model sizes, and dataset scales in LLM pretraining \citep{ScalingLaws, ScalingLanguageModels}. However, in practical scenarios where dataset size is limited, or data mixtures vary, scaling laws exhibit significant variations \citep{Hoffmann}. Several studies have extended scaling laws to account for these complexities. \citep{Muennighoff, RepeatedScaling} explore the impact of data repetition levels on scaling behaviors, while \citep{BiMix} investigate scaling dynamics under different domain proportions and dataset sizes. To optimize data compositions, \citep{regmix} propose a regression model for predicting optimal mixtures, and \citep{AutoScale} further analyze optimal compositions across varying scales. Additionally, \citep{D-CPT} focus on identifying the best data mixtures for the continued pretraining stage, providing insights into refining pretraining strategies under diverse constraints.
\section{Methodology}

Our approach can be illustrated in 4 parts: 1) We propose the QuaDMix framework, which utilizes a unified parameterized function to govern the data sampling process. 2) We conduct small-scale experiments to explore how different parameter settings within QuaDMix affect the performance of LLM. 3) We train a regression model to capture these effects, using it to identify the optimal parameters. 4) With the optimal parameter settings, we sample large-scale data and train a large language model.


\subsection{Design of QuaDMix}
\label{3.1}

We design QuaDMix as a sampling algorithm that simultaneously accounts for data quality and diversity, as shown in Figure \ref{fig:image_label}.
Given a pretraining dataset $X$, we define a sampling function $ S(x,\bm{q}_x,d_x;\bm{\theta}) $, which determines the expected sampling times of each data point $x$ based on its data feature $\bm{q}_x$ and $d_x$. Here $\bm{q_x}$ represents the quality score vector, which includes multiple quality criteria, and $d_x$ denotes the domain to which $x$ belongs. $\bm{\theta}$ are the parameters to be optimized.
The output of this function is fractional value, e.g. $a.b$, meaning the document will be sampled for $a$ times plus another random sampling with probability $b$.
\\
\noindent\textbf{Feature Extraction} To measure a sample's contribution to diversity and its quality, we propose using domain classification and $N$ quality scorers to label the pretraining data. 
Specifically, we use a domain classifier to divide the dataset into $M$ domains, where $x$ will be assigned a domain label $d_x$. Then we use $N$ quality scorers to compute the quality vector $\bm{q_x}=(q_{1,x},...,q_{N,x})$, 
and for each $q_{n,x}$, 
a smaller value indicates a better quality on that dimension. For the sake of simplicity, we omit $x$ 
in the subscript in the following discussion.

\noindent\textbf{Quality Ranking} We first define a merging function that integrates the scores from various quality filters, aiming to incorporate complementary information provided by different criteria. Assuming there are $N$ criteria, for any individual example $\bm{x}$ belonging to domain $m$, the merged quality score is calculated by 

\begin{equation}
\bar{q}=\sum_{n=1}^N \sigma(q_{n})\alpha_{n,m},
\end{equation}

where $\bm{\alpha}_m$ are the merging parameters for domain $m$. We utilize separate merging parameters to balance the quality criteria across different domains, as the criteria exhibit varying preferences depending on the domain. $\sigma$ is a normalization function to align the scales of quality criteria.

We then sort the data based on the merged quality score. The sorting is operated separately in each domain. The merged quality rank $\bar{r}$ is calculated by computing the percentile of the data within that domain. That is 
\begin{equation}
\bar{r}=\frac{|\{x|d_x=m,\bar{q}_x<=\bar{q}\}|}{|\{x|d_x=m\}|}.
\end{equation}
Here we calculate the size of the set by adding up the number of tokens for all sample within the set. For a given example in domain $m$ with $\bar{r}=0.05$, this means that 95\% of the tokens in that domain have a worse quality compared to this example. (Note that we use smaller quality scores to represent higher quality.)
\\

\noindent\textbf{Quality Sampling} Next, we define the sampling function. We take the assumption that higher-quality data should be sampled more frequently in the final dataset. This assumption is supported by evidence \citep{fineweb-edu}, which demonstrates that applying a higher quality threshold improves downstream performance. For any example in domain $m$ with merged quality rank $\bar{r}$, the value of the sampling function is determined by 

\begin{equation} 
S(\bar{r})=\left\{
\begin{array}{lr}
(\frac{2}{1+e^{-\lambda_m(\omega_m-\bar{r})}})^{\eta_m}+\epsilon_m,& \bar{r}<=\omega_m \\
\epsilon_m,& \bar{r}>\omega_m
\end{array}
\right.
\end{equation}

We denote $\bm{\beta}_m=(\lambda_m,\omega_m,\eta_m,\epsilon_m)$ as the sampling parameters for domain $m$. We use a format of sigmoid to ensure the sampling value is monotonically decreasing as the quality rank goes up (worse quality) and $\lambda_m$ is used to adjust how fast it decreases. $\omega_m$ controls the quality percentile threshold, determining the minimum quality level we aim to retain. $\eta_m$ is a scaling parameter that adjusts the sampling values, while $\epsilon_m$ introduces randomness to incorporate data from all quality ranges. By applying different sampling parameters across domains, we achieve flexible control over domain proportions.

In summary, by integrating (1),(2), and (3), we define the sampling function for individual domain $m$, with the parameters structured as $\bm{\theta}_m=(\bm{\alpha}_m,\bm{\beta}_m)$. The total number of parameters is $(N+4)\times M$, where $N$ represents the number of used quality criteria and $M$ denotes the total number of distinct domains.

\subsection{Proxy Model Experiments}
\label{3.2}
We first sample a set of values for each parameter defined above, subsequently generating corresponding datasets using the QuaDMix sampling function. Following this, a series of small proxy models are trained on each dataset and evaluated on the validation set to compute the validation loss.
\\
\noindent\textbf{Parameter Sampling} The parameter space requires careful design to encompass valuable regions, while avoiding extreme conditions. We sample from the parameter space as following:
\begin{algorithm}
    \caption{Parameter Sampling for QuaDMix}
    \begin{algorithmic}
        \ENSURE{$\bm{\theta}$}
        \REQUIRE{$N,M$}
        \STATE {Sample $(a_1,...,a_N)\sim U(0,1)$}
        \STATE {$\tilde{a}_n=\frac{a_n}{\sum_i a_i}$}
        \FOR{$m=1$ \TO $M$ } 
        \STATE {Sample $(b_{1,m},...,b_{N,m})\sim U(0,1)$}
        \STATE {$\tilde{b}_{n,m}=\frac{\tilde{a}_n b_{n,m}}{\sum_i \tilde{a}_i b_{i,m}}$}
        \STATE {$\bm{\alpha}_m=(\tilde{b}_{n,m}),n=1,...,N$}
        \STATE {Sample $(\lambda_m,\omega_m,\eta_m,\epsilon_m)\sim U(0,1)$}
        \STATE {$\tilde{\lambda}_m=10^{3\lambda_m},\ \tilde{\omega}_m=0.1\omega_m$}
        \STATE {$\tilde{\eta}_m=\eta_m,\ \tilde{\epsilon}_m=\epsilon_m/1000$}
        \STATE {$\bm{\beta}_m=(\tilde{\lambda}_m,\tilde{\omega}_m,\tilde{\eta}_m,\tilde{\epsilon}_m)$}
        \STATE {$\bm{\theta}_m=(\bm{\alpha}_m,\bm{\beta}_m)$}
        \ENDFOR
        \STATE {$\bm{\theta} = (\bm{\theta}_1,...,\bm{\theta}_M)$}
    \end{algorithmic}
\end{algorithm}

In the algorithm above, we introduce a global weight for each quality criteria, with the final weight computed by multiplying the global weight by the domain-specific weight. Without this global weight, the expected average weight across domains for each quality criterion would always be $1/N$, which fails to account for the scenario where one quality criterion may suppress another overall. For $\bm{\beta}_m$, we rescale them accordingly to ensure domain proportions and quality thresholds remain within a reasonable range. Using this process, we generate 3,000 sets of parameters $\bm{\theta}_i$ and then sample with QuaDMix from our training data, producing 3,000 proxy datasets, denoting as $D_i$.

\noindent\textbf{Proxy Model Training}
Next we train the proxy models on each proxy datasets from scratch.
$$f_i^*=\arg\min _{f} L(f,D_i)$$
After training, we evaluate the proxy models by calculating the loss on the target evaluation datasets. 
$$L_i=L(f_i^*,D_{eval})$$

\subsection{Parameter Optimizing}
\noindent\textbf{Regression Model Fitting}
The next step is to determine the correlation between the sampled QuaDMix parameters and model performance. We formulate this as a regression problem, as proposed in \citep{regmix}, with the goal of learning a function that predicts the target value based on the input features. Specifically, we optimize a regressor $R$ with

$$R^*=\arg\min_R \sum_i||R(\bm{\theta}_i)-L_i||^2$$

We evaluate different types of regressors and select LightGBM \citep{lightgbm}, which ensembles multiple decision trees, to predict the target value.


\noindent\textbf{Optimal Parameter Estimation}
Once the regressor is trained, we search the input space to find the optimal parameters that minimize the predicted loss. Rather than performing a random search across the entire space, we sample 100,000 data points using the algorithm outlined in Section \ref{3.2} to mitigate the influence of outliers on the regressor. To further reduce the variance in the regression predictions, we sort the data points based on their predicted target values and calculate the average of the top 10 data points to determine the final output.

\subsection{Large-scale Model Experiments}
We then use the optimal parameters to generate large-scale datasets for training large-scale models. In practice, since sorting the quality scores across the entire dataset is computationally expensive, we estimate the quality percentile by randomly selecting a subset of 10,000 documents. Within this subset, we calculate the mapping between the quality percentile and quality score, and then apply this mapping to the entire dataset.

\begin{table*}[h]
    \centering
    \begin{tabular}{l|c|c|c|c|c|c}
        \hline
        
         & Selected & Reading & Commonsense &  &  &  \\  
        Methods & Token & Comprehension & Reasoning & Knowledge & Math & Average \\  
        
        \hline
        \hline
        Random Selection & 500B & 32.9	& 51.6 & 17.4 & 2.8 & 32.3   \\ 
        DSIR & 72B & 34.9 &	49.2 & 17.5 &	6.9 & 32.7   \\ 
        RegMix & 500B & 35.5	& 52.4 & 17.7 & 3.5 &33.6  \\
        Fineweb-edu & 30B & 41.4 & 55.5 & 20.1 & 6.0 & 37.4  \\  
        AskLLM & 30B & 38.9 &	54.2 &	19.0 &	2.3 & 35.5  \\  
        DCLM & 30B & 41.2 &	53.1 &	19.8 & 8.2 & 36.7  \\  
        Criteria Mix & 74B & 40.1	& 53.7 & 20.0 & 3.1 &36.0  \\
        
        \hline
        \hline
        QuaDMix-OH & 30B & 44.0 & 55.7 & 21.0 &	10.2 & 39.0  \\ 
        QuaDMix-BMK & 30B & \textbf{44.8} &	\textbf{55.7} & \textbf{21.3} & \textbf{11.5}	& \textbf{39.5}  \\ 
        \hline
    \end{tabular}
    \caption{QuaDMix outperforms the methods focusing only on data quality or data mixture. With benchmark training set as the target, the results further boost. }
    \label{tab:results}
\end{table*}

\section{Experiments on Regression Model}

\subsection{Experiment Setup}
\noindent\textbf{Datasets} We conduct our experiment on RefinedWeb \citep{refinedweb}. It is an English large-scale dataset for the pretraining of large language models and consists of over 570B(billion) tokens. For the small proxy datasets, we sample it from a subset of RefinedWeb, each containing 1B tokens.


\noindent\textbf{Feature Extraction} We generate the necessary data features including data quality and domain index with 3 individual quality filters, AskLLM \citep{askllm}, Fineweb-Edu \citep{fineweb-edu}, DCLM \citep{dclm} and 1 domain classifier \citep{NeMo}, which classify the data into 26 different domains with a Deberta V3 \citep{DeBERTaV3} architecture. 

\noindent\textbf{Training and evaluation} For the proxy models, we train them on the proxy datasets for 1B tokens, taking 1 NVIDIA H100 GPU hour and calculate the loss on the validation datasets. To construct the validation datasets, we sample from the instruction-formatted dataset OpenHermes 2.5 \citep{Openhermes}. As demonstrated in \citep{dclm}, this dataset is used to train a robust quality filter. To improve efficiency, we sampled 10k samples from it to form a validation subset, named openhermes-10k. Additionally, we test on the training data from the downstream tasks including HellaSwag, ARC-E, ARC-C, MMLU, and TriviaQA to demonstrate the model's ability to optimize for specific downstream tasks by altering the target evaluation datasets.

For the regression model, we split the data into 2800/200 for training and validation. We use Mean Absolute Error (MAE) as the evaluation metric, which calculates the average absolute differences between predicted and actual values.

\begin{figure}[t] 
    \centering
    \includegraphics[width=0.5\textwidth]{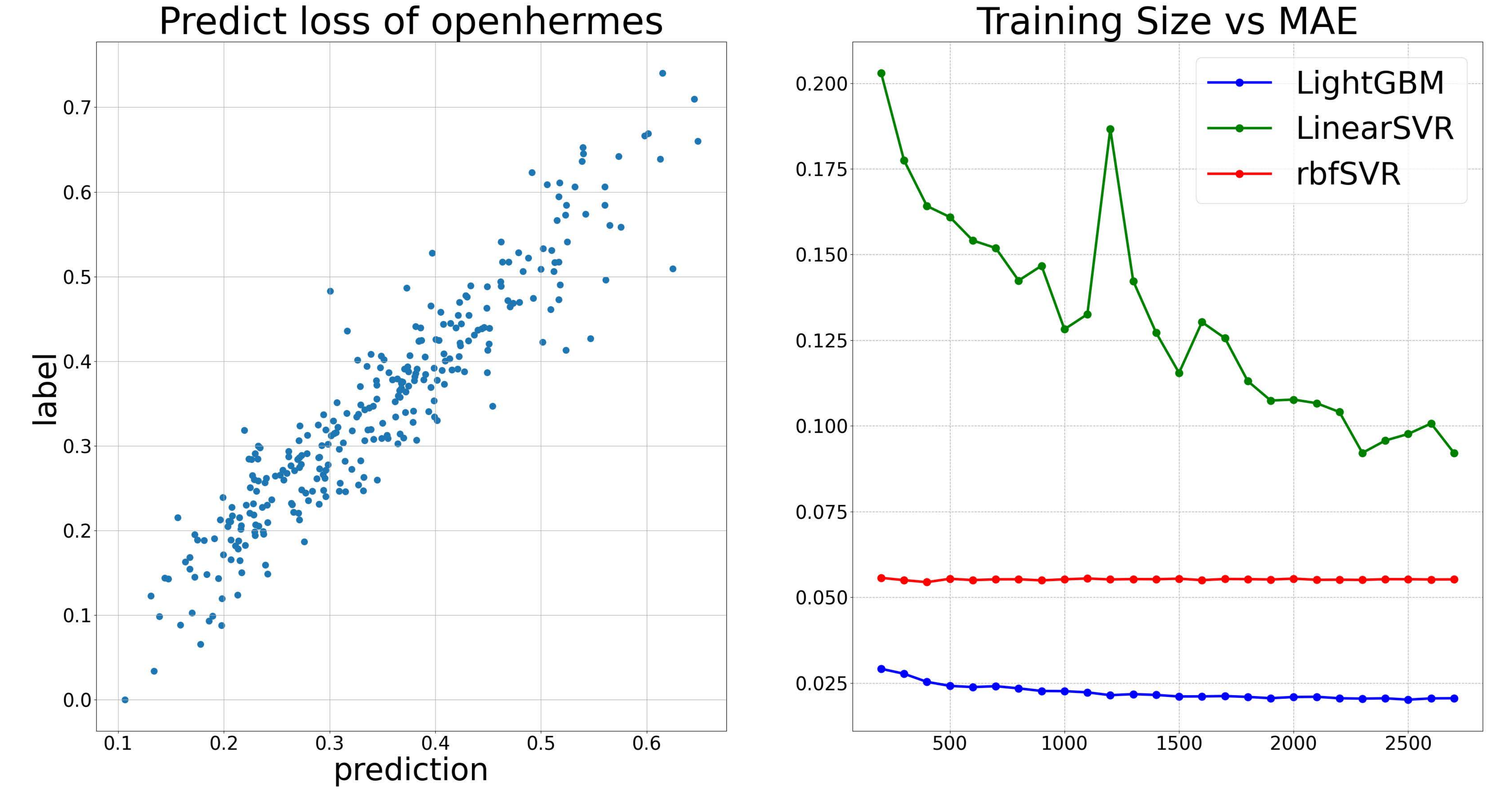} 
    \caption{Left: The prediction model loss vs real model loss. Right: The regression model performance (MAE) vs training size. }
    \label{fig:training_size_vs_mae} 
\end{figure}

\subsection{Results}
We show the results of regression models in Figure \ref{fig:training_size_vs_mae}. The left figure shows strong correlation between the predicted loss and the real model loss (evaluated on OpenHermes) on the validation set, providing the evidence that there exists statistical pattern between the QuaDMix parameters and the model performance. We compare three regression models in the right figure. We can see LightGBM yields better accuracy in predicting the model performance than SVR \citep{SVR} with Linear kernel and RBF kernel. Also, with larger training size, the accuracy keeps increasing. Considering the training budget, we conduct 3000 proxy experiments in total to get a better results.

\section{Experiments on Language Model}

\begin{table*}[h]
    \centering
    \begin{tabular}{c|c|c|c|c|c|c|c|c}
        \hline
         &&& Selected & Reading & Commonsense &  &  &  \\
        A&F& D & Token & Comprehension & Reasoning & Knowledge & Math & Average \\  
        
        \hline
        \hline
        \checkmark &&& 30B & 38.9 &	53.5 & 18.6 & 2.9 &	35.2  \\ 
         &\checkmark&& 30B & 41.4 &	55.5 &	20.1 &	6.0	& 37.4   \\ 
         &&\checkmark& 30B & 41.3 & 53.4 &	19.7 &	\textbf{12.2} & 37.3  \\  
         \checkmark&\checkmark&& 30B & 41.9 &	55.6 &	20.0 & 5.1 & 37.5   \\ 
         \checkmark&&\checkmark& 30B & 41.8 &	54.6 &	19.8 &	9.1 &	37.5  \\ 
         &\checkmark&\checkmark& 30B & 43.5	& 55.6	& 20.8 &	9.6 & 38.7  \\  
        \hline
        \hline
        \checkmark&\checkmark&\checkmark& 90B & 40.7 &	55.2 &	19.5 & 4.6 & 36.8   \\  
        \checkmark&\checkmark&\checkmark& 180B 	 & 37.8 & 53.9 & 18.9 & 2.8 &	35.1   \\ 
        \hline
        \hline
        \checkmark&\checkmark&\checkmark& 30B & \textbf{44.0} & \textbf{55.7} & \textbf{21.0} &	10.2 & \textbf{39.0}  \\ 
       
        \hline
    \end{tabular}
    \caption{QuaDMix-OH with different settings on quality filters (AskLLM (A), Fineweb-edu (F), DCLM (D)) and selected tokens. }
    \label{tab:ablation}
\end{table*}

\begin{figure*}[h]
\centering
\begin{subfigure}{.45\textwidth}
    \centering
    \includegraphics[width=1.0\textwidth]{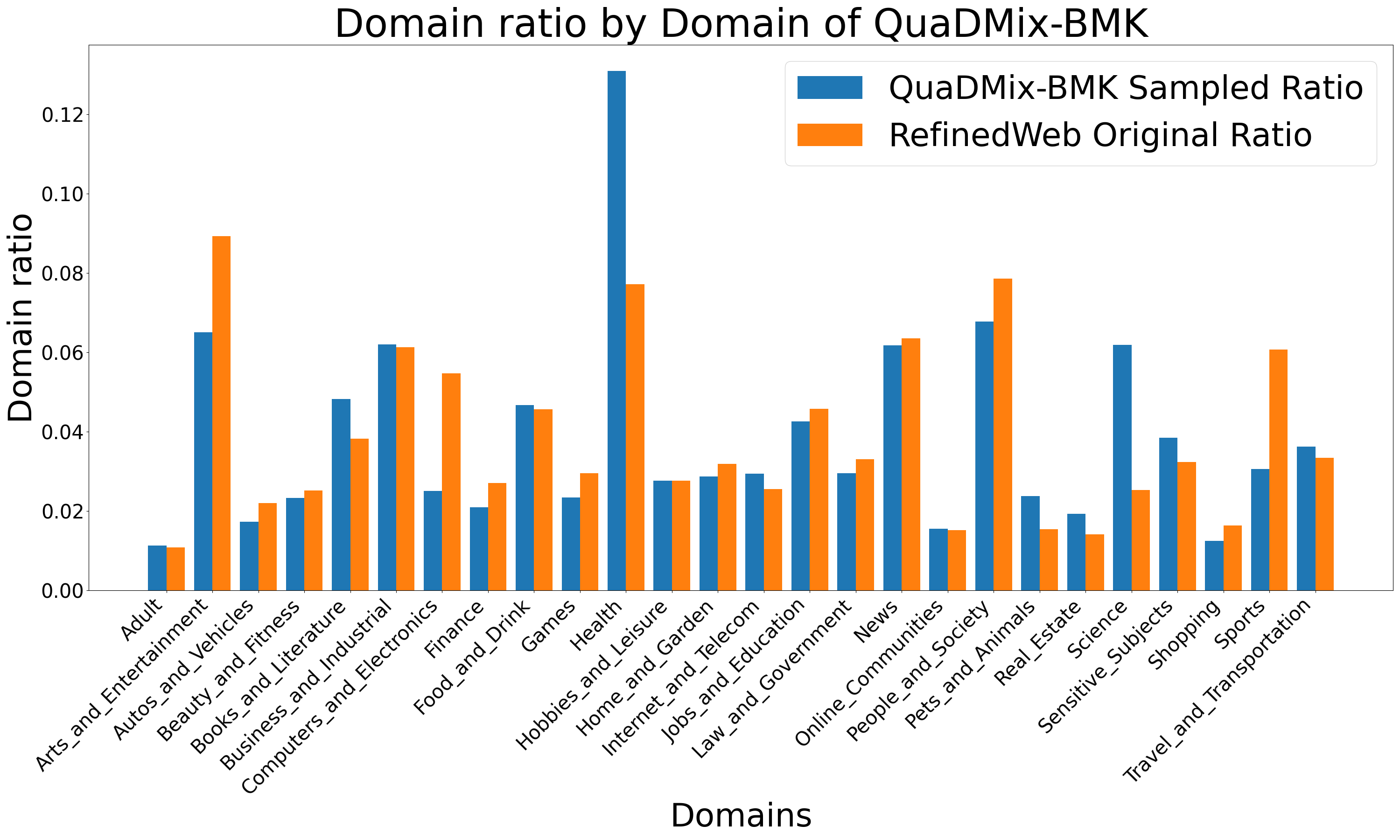}  

    \label{fig:domain_ratio_bmk}
\end{subfigure}
\begin{subfigure}{.45\textwidth}
    \centering
    \includegraphics[width=1.0\textwidth]{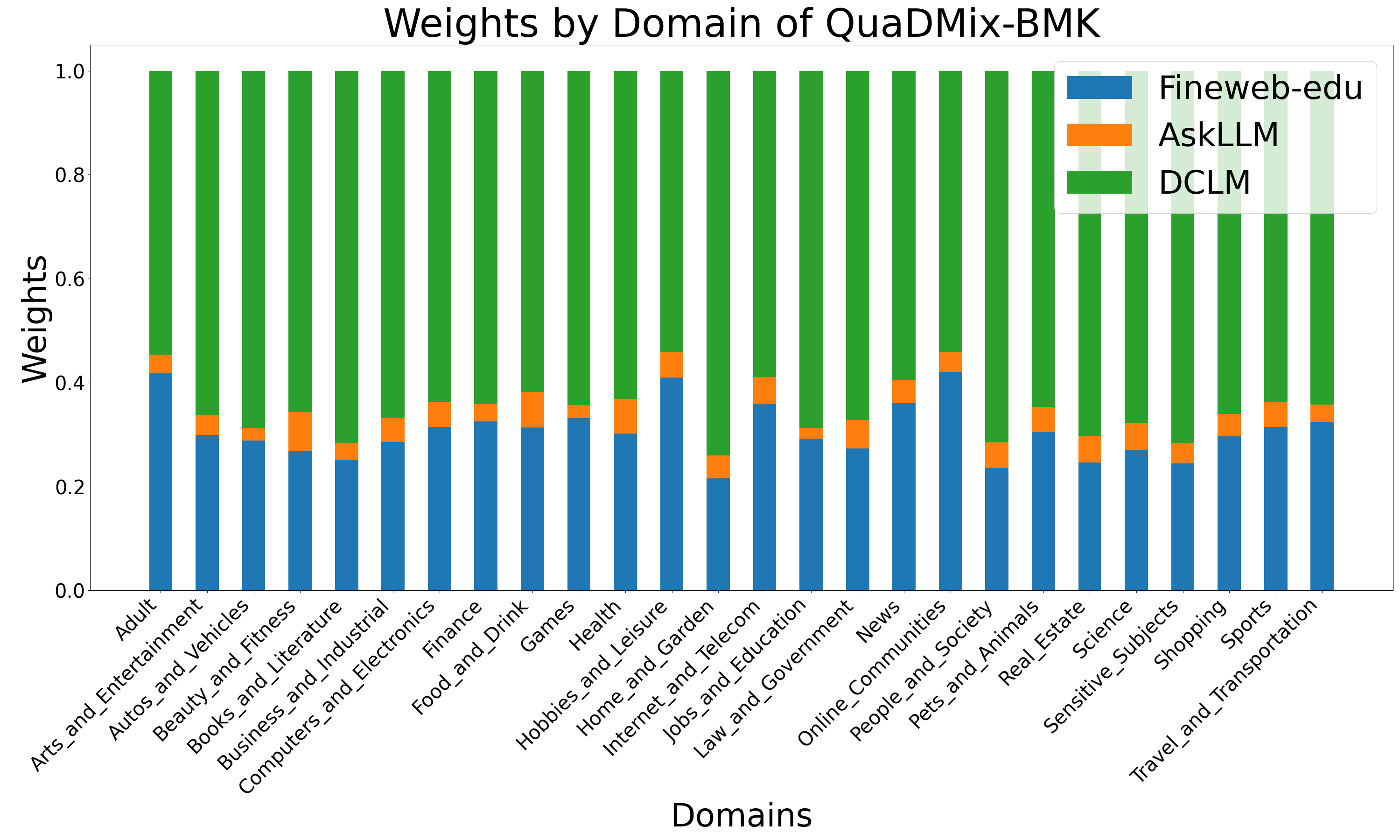}  

    \label{fig:merge_weights_bmk}
\end{subfigure}
\caption{The visualization of optimal parameters from QuaDMix-BMK}
\label{fig:vis_bmk}
\end{figure*}
In this section we compare different methods of data selection and mixture with QuaDMix by training language models from scratch and evaluating on various downstream tasks. 

\subsection{Experiment Setup}
\noindent\textbf{Training and evaluation} We train the language model with 530M parameters from scratch for 500B tokens, taking 32 NVIDIA GPU for 3 days. We use transformer architecture \citep{Attention}, SwiGLU \citep{SwiGLU} as the activation function and RoPE embeddings \citep{ROPE}. 
Then we evaluate the model performance using lm-eval-harness \citep{harness}. We choose $9$ downstream tasks, including 3 commonsense reasoning tasks (PIQA \citep{PIQA}, HellaSwag \citep{hellaswag}, OpenBookQA \citep{OpenBookQA}), 3 reading comprehension tasks (ARC-E/C \citep{ARC}, Triviaqa \citep{triviaqa}), 1 math problem solving task (SVAMP \citep{SVAMP}) and 2 knowledge intensive tasks (MMLU \citep{MMLU}, NQ-open \citep{NQ,NQ_open}). For each benchmark, we used normalized accuracy as the evaluation metric. Some modifications on the testing logic are applied for numerical stability. 

\subsection{Data Selection Methods}
We generate the training data from the RefinedWeb dataset using following data selection methods.

\noindent \textbullet\ \textbf{Random Selection}: Documents are randomly selected from the whole dataset. 

\noindent \textbullet\ \textbf{Fineweb-edu Classifier}: Documents are scored with Fineweb-edu Classifier \citep{fineweb-edu} with top-k selection

\noindent \textbullet\ \textbf{AskLLM}: Documents are scored with the probability of generating "Yes" from a prompted large language model \citep{askllm}. The top-k documents are selected.

\noindent \textbullet\ \textbf{DCLM}: Documents are scored with fasttext based classifier \citep{dclm} with top-k selection.

\noindent \textbullet\ \textbf{Criteria Mix}: Following \citep{QuRating}, the selected data from the above three filters are merged, with duplicated documents removed.

\noindent \textbullet\ \textbf{DSIR}: Documents are sampled based on the importance calculated with the N-gram features \citep{DSIR}.

\noindent \textbullet\ \textbf{RegMix}: Following \citep{regmix}, we conduct 512 1M porxy experiments and randomly select data using the optimized data mixtures.

\noindent \textbullet\ \textbf{QuaDMix-OH}: Documents are sampled with the proposed QuaDMix, where Openhermes is used as the validation set for the proxy experiments

\noindent \textbullet\ \textbf{QuaDMix-BMK}: Documents are sampled with the proposed QuaDMix, where the training set of 5 downstream tasks (HellaSwag, ARC-E, ARC-C, MMLU, TriviaQA) are used as the validation set to generate the optimal QuaDMix parameters.

\begin{figure*}[h] 
    \centering
    \includegraphics[width=1.0\textwidth]{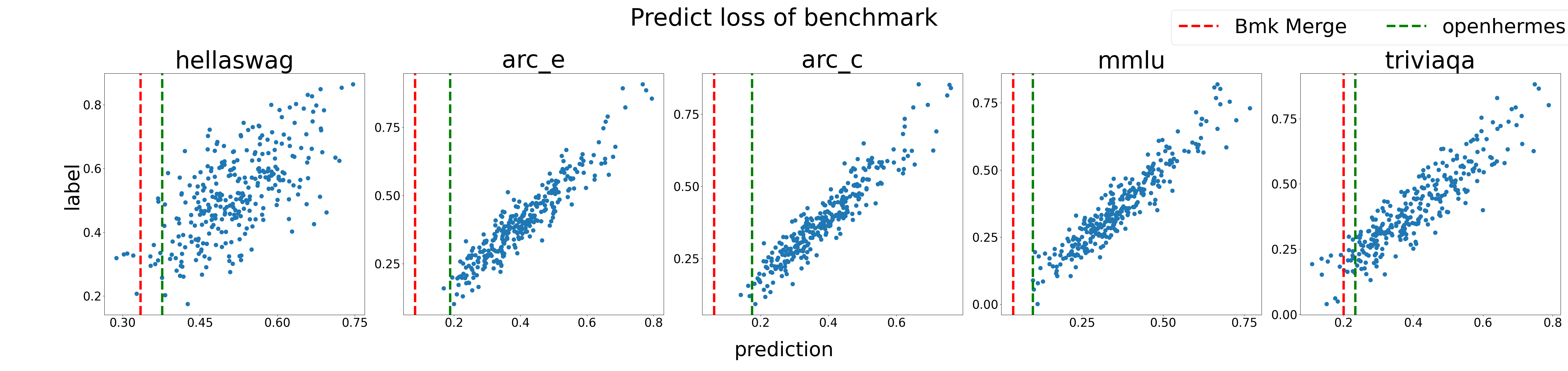} 
    \caption{The prediction loss of QuaDMix-BMK surpasses QuaDMix-OH on all 5 downstream tasks.}
    \label{fig:predict_loss} 
\end{figure*}

\begin{table*}[h]
    \centering
    \begin{tabular}{c|c|c|c|c|c}
        \hline
        Method & HellaSwag & ARC-C & ARC-E & MMLU & TriviaQA \\  
        \hline
        \hline
        QuaDMix-OH  & \textbf{56.5} & 39.2 & 71.1 & 34.1 &	21.6   \\ 
        \hline
        QuaDMix-BMK  & 56.1 & \textbf{40.2} & \textbf{71.3} & \textbf{34.4} &	\textbf{22.8}   \\ 
        \hline
    \end{tabular}
    \caption{QuaDMix-OH vs QuaDMix-BMK on 5 downstream tasks. The trend mostly agree with the prediction loss on proxy model except for HellaSwag.}
    \label{tab:bmk}
\end{table*}

\subsection{Results}
The results are summerized in Table \ref{tab:results}. We can see that QuaDMix outperforms the methods focusing only on data quality or data mixture on all the benchmarks, proving the necessity of jointly considering quality and diversity. It also shows that the proxy model experiments can well indicate the performance on large scale model. With loss of the benchmark training set as the target when training the regression model, the results further boost. This prove the ability of QuaDMix of optimizing for specific downstream tasks by choosing evaluation datasets in proxy experiments which are more related to the downstream tasks.

\noindent\textbf{Analysis of optimal QuaDMix parameters} We show the optimal data mixtures and merging parameters of quality filters from QuaDMix-BMK in Figure \ref{fig:vis_bmk}. We see that the Health and Science domain are upsampled for large margin, while Sports and Computers downsampled, indicating that the downstream tasks we choose have preference for specific domains. The right figure shows that the DCLM quality filter contributes most to the merged quality score, while AskLLM only occupies a small weight among the three filters.

\section{Ablations}

\noindent\textbf{Quality Merging Benefits Selection} To prove the necessity of quality score merging, we select different combinations of quality filters by manually setting the weight of certain filters to $0$ when finding the optimal QuaDMix parameters. As shown in Table \ref{tab:ablation}, merging with all three quality filters shows the best performance. Although using one quality filter can be optimal for one specific task, for example DCLM-only for MATH, the merging process reduces intrinsic bias within the quality filters and outperforms in general ability, which is essential for language model pretraining.

\noindent\textbf{More Tokens not always good} We also experiment with selecting more tokens by loosing the sampling parameter $\omega$ in QuaDMix. In that way we introduce more diversed tokens but lower quality into the training. The results in Table \ref{tab:ablation} show that selecting 30B
 tokens, i.e. documents with top5\% quality yields the best result, meaning that curing data quality contributes more than increasing the number of unique tokens within this range.

\noindent\textbf{Proxy Ability of Small Models} How well the prediction loss on proxy models forecasts the performance on large-scale models is the key factor of QuaDMix. To study this, we train 5 separate regression models, each using the loss on training set of one benchmark as the target. The results on the validation set are shown as blue points in Figure \ref{fig:predict_loss}. We notice that HellaSwag has larger variance than others, which indicates there may be more influencing factors related with HellaSwag, making the loss on it harder to predict. Then we predict the loss for optimal parameters from QuaDMix-OH and QuaDMix-BMK using each regression model as shown in Figure \ref{fig:predict_loss}. It is reasonable to see the loss of QuaDMix-BMK surpasses QuaDMix-OH on all tasks since QuaDMix-BMK utilizes benchmark training set as optimizing target. Finally we report the performance of large model in Table \ref{tab:bmk}. Except for HellaSwag, QuaDMix-BMK outperforms QuaDMix-OH on other tasks, which agrees with the trend on prediction loss. The inconsistent conclusion on HellaSwag is because the predict loss has larger variance as mentioned above, making the proxy ability lower than other tasks. How to further increase the proxy ability is one of the future direction to explore.

\section{Conclusion}
In this paper, we propose a novel data selection method QuaDMix that jointly optimizes the data quality and diversity for language model pretraining. We design a parameterized space that controls both the data quality and diversity, and conduct proxy experiments to find the correlation between the parameter and model performance. The training data generated with optimal parameters are proved to outperform others on various downstream tasks. 
\section{Limitations}
We note several limitations of our work. There exist improvement space for the design of parameter space of QuaDMix. For example the parameters of sampling function may generate similar functions under different parameters, which will cause redundancy and introduce uncertainty into the regression model. Secondly, the searching in the parameter space for optimal parameters is inefficient. We use random guessing in a space with 200 more dimensions, for certain the current optimal parameter is a local minimum and how to effectively search in the parameter space remains unclear. Finally, the proxy ability of small models is crucial, what is the systematic way to improve it is an important yet less explored topic. However, QuaDMix provides a useful solution for jointly optimize for data quality and diversity, and it worth continually exploring on the limitations mentioned above.

\bibliography{acl_latex}

\begin{thebibliography}{58}
\providecommand{\natexlab}[1]{#1}

\bibitem[{Abbas et~al.(2023)Abbas, Tirumala, Simig, Ganguli, and Morcos}]{SemDeDup}
Amro Abbas, Kushal Tirumala, Dániel Simig, Surya Ganguli, and Ari~S. Morcos. 2023.
\newblock \href {https://arxiv.org/abs/2303.09540} {Semdedup: Data-efficient learning at web-scale through semantic deduplication}.
\newblock \emph{Preprint}, arXiv:2303.09540.

\bibitem[{Bai et~al.(2024)Bai, Yang, Wong, Peng, Zhuang, Zhang, Wu, Qiu, Zhang, Yuan, and He}]{Multi-Agent}
Tianyi Bai, Ling Yang, Zhen~Hao Wong, Jiahui Peng, Xinlin Zhuang, Chi Zhang, Lijun Wu, Jiantao Qiu, Wentao Zhang, Binhang Yuan, and Conghui He. 2024.
\newblock \href {https://arxiv.org/abs/2410.08102} {Multi-agent collaborative data selection for efficient llm pretraining}.
\newblock \emph{Preprint}, arXiv:2410.08102.

\bibitem[{Bisk et~al.(2019)Bisk, Zellers, Bras, Gao, and Choi}]{PIQA}
Yonatan Bisk, Rowan Zellers, Ronan~Le Bras, Jianfeng Gao, and Yejin Choi. 2019.
\newblock \href {https://api.semanticscholar.org/CorpusID:208290939} {Piqa: Reasoning about physical commonsense in natural language}.
\newblock In \emph{AAAI Conference on Artificial Intelligence}.

\bibitem[{Brown et~al.(2020)Brown, Mann, Ryder, Subbiah, and et~al.}]{Brown:2023}
Tom Brown, Benjamin Mann, Nick Ryder, Melanie Subbiah, and et~al. 2020.
\newblock \href {https://proceedings.neurips.cc/paper_files/paper/2020/file/1457c0d6bfcb4967418bfb8ac142f64a-Paper.pdf} {Language models are few-shot learners}.
\newblock In \emph{Advances in Neural Information Processing Systems}, volume~33, pages 1877--1901.

\bibitem[{Cheng et~al.(2024)Cheng, Gu, Huang, Bi, Huang, and Wei}]{InstructionPT}
Daixuan Cheng, Yuxian Gu, Shaohan Huang, Junyu Bi, Minlie Huang, and Furu Wei. 2024.
\newblock \href {https://doi.org/10.18653/v1/2024.emnlp-main.148} {Instruction pre-training: Language models are supervised multitask learners}.
\newblock In \emph{Proceedings of the 2024 Conference on Empirical Methods in Natural Language Processing}, pages 2529--2550, Miami, Florida, USA. Association for Computational Linguistics.

\bibitem[{Chowdhery et~al.(2023)Chowdhery, Narang, Devlin, Bosma, Mishra, and et~al.}]{Chowdhery:2023}
Aakanksha Chowdhery, Sharan Narang, Jacob Devlin, Maarten Bosma, Gaurav Mishra, and et~al. 2023.
\newblock \href {http://jmlr.org/papers/v24/22-1144.html} {Palm: Scaling language modeling with pathways}.
\newblock \emph{Journal of Machine Learning Research}, 24(240):1--113.

\bibitem[{Clark et~al.(2018)Clark, Cowhey, Etzioni, Khot, Sabharwal, Schoenick, and Tafjord}]{ARC}
Peter Clark, Isaac Cowhey, Oren Etzioni, Tushar Khot, Ashish Sabharwal, Carissa Schoenick, and Oyvind Tafjord. 2018.
\newblock \href {https://arxiv.org/abs/1803.05457} {Think you have solved question answering? try arc, the ai2 reasoning challenge}.
\newblock \emph{Preprint}, arXiv:1803.05457.

\bibitem[{Drucker et~al.(1996)Drucker, Burges, Kaufman, Smola, and Vapnik}]{SVR}
Harris Drucker, Christopher J.~C. Burges, Linda Kaufman, Alex Smola, and Vladimir Vapnik. 1996.
\newblock \href {https://proceedings.neurips.cc/paper_files/paper/1996/file/d38901788c533e8286cb6400b40b386d-Paper.pdf} {Support vector regression machines}.
\newblock In \emph{Advances in Neural Information Processing Systems}, volume~9. MIT Press.

\bibitem[{Du et~al.(2022)Du, Huang, Dai, Tong, Lepikhin, Xu, Krikun, Zhou, Yu, Firat, Zoph, Fedus, Bosma, Zhou, Wang, Wang, Webster, Pellat, Robinson, Meier-Hellstern, Duke, Dixon, Zhang, Le, Wu, Chen, and Cui}]{glam}
Nan Du, Yanping Huang, Andrew~M Dai, Simon Tong, Dmitry Lepikhin, Yuanzhong Xu, Maxim Krikun, Yanqi Zhou, Adams~Wei Yu, Orhan Firat, Barret Zoph, Liam Fedus, Maarten~P Bosma, Zongwei Zhou, Tao Wang, Emma Wang, Kellie Webster, Marie Pellat, Kevin Robinson, Kathleen Meier-Hellstern, Toju Duke, Lucas Dixon, Kun Zhang, Quoc Le, Yonghui Wu, Zhifeng Chen, and Claire Cui. 2022.
\newblock \href {https://proceedings.mlr.press/v162/du22c.html} {{GL}a{M}: Efficient scaling of language models with mixture-of-experts}.
\newblock In \emph{Proceedings of the 39th International Conference on Machine Learning}, volume 162 of \emph{Proceedings of Machine Learning Research}, pages 5547--5569. PMLR.

\bibitem[{Fan et~al.(2024)Fan, Pagliardini, and Jaggi}]{DoGE}
Simin Fan, Matteo Pagliardini, and Martin Jaggi. 2024.
\newblock \href {https://arxiv.org/abs/2310.15393} {Doge: Domain reweighting with generalization estimation}.
\newblock \emph{Preprint}, arXiv:2310.15393.

\bibitem[{Gao et~al.(2020)Gao, Biderman, Black, Golding, Hoppe, Foster, Phang, He, Thite, Nabeshima, Presser, and Leahy}]{pilecc}
Leo Gao, Stella Biderman, Sid Black, Laurence Golding, Travis Hoppe, Charles Foster, Jason Phang, Horace He, Anish Thite, Noa Nabeshima, Shawn Presser, and Connor Leahy. 2020.
\newblock \href {https://arxiv.org/abs/2101.00027} {The pile: An 800gb dataset of diverse text for language modeling}.
\newblock \emph{Preprint}, arXiv:2101.00027.

\bibitem[{Gao et~al.(2023)Gao, Tow, Abbasi, Biderman, Black, DiPofi, Foster, Golding, Hsu, Le~Noac'h, Li, McDonell, Muennighoff, Ociepa, Phang, Reynolds, Schoelkopf, Skowron, Sutawika, Tang, Thite, Wang, Wang, and Zou}]{harness}
Leo Gao, Jonathan Tow, Baber Abbasi, Stella Biderman, Sid Black, Anthony DiPofi, Charles Foster, Laurence Golding, Jeffrey Hsu, Alain Le~Noac'h, Haonan Li, Kyle McDonell, Niklas Muennighoff, Chris Ociepa, Jason Phang, Laria Reynolds, Hailey Schoelkopf, Aviya Skowron, Lintang Sutawika, Eric Tang, Anish Thite, Ben Wang, Kevin Wang, and Andy Zou. 2023.
\newblock \href {https://doi.org/10.5281/zenodo.10256836} {A framework for few-shot language model evaluation}.

\bibitem[{Ge et~al.(2024)Ge, Ma, Chen, Li, and Ding}]{BiMix}
Ce~Ge, Zhijian Ma, Daoyuan Chen, Yaliang Li, and Bolin Ding. 2024.
\newblock \href {https://arxiv.org/abs/2405.14908} {Bimix: Bivariate data mixing law for language model pretraining}.
\newblock \emph{Preprint}, arXiv:2405.14908.

\bibitem[{Gu et~al.(2024)Gu, Dong, Wang, Hao, Dong, Wei, and Huang}]{OptimalControl}
Yuxian Gu, Li~Dong, Hongning Wang, Yaru Hao, Qingxiu Dong, Furu Wei, and Minlie Huang. 2024.
\newblock \href {https://arxiv.org/abs/2410.07064} {Data selection via optimal control for language models}.
\newblock \emph{Preprint}, arXiv:2410.07064.

\bibitem[{Gunasekar et~al.(2023)Gunasekar, Zhang, Aneja, Mendes, Giorno, Gopi, Javaheripi, Kauffmann, de~Rosa, Saarikivi, Salim, Shah, Behl, Wang, Bubeck, Eldan, Kalai, Lee, and Li}]{phi1}
Suriya Gunasekar, Yi~Zhang, Jyoti Aneja, Caio César~Teodoro Mendes, Allie~Del Giorno, Sivakanth Gopi, Mojan Javaheripi, Piero Kauffmann, Gustavo de~Rosa, Olli Saarikivi, Adil Salim, Shital Shah, Harkirat~Singh Behl, Xin Wang, Sébastien Bubeck, Ronen Eldan, Adam~Tauman Kalai, Yin~Tat Lee, and Yuanzhi Li. 2023.
\newblock \href {https://arxiv.org/abs/2306.11644} {Textbooks are all you need}.
\newblock \emph{Preprint}, arXiv:2306.11644.

\bibitem[{He et~al.(2023)He, Gao, and Chen}]{DeBERTaV3}
Pengcheng He, Jianfeng Gao, and Weizhu Chen. 2023.
\newblock \href {https://arxiv.org/abs/2111.09543} {Debertav3: Improving deberta using electra-style pre-training with gradient-disentangled embedding sharing}.
\newblock \emph{Preprint}, arXiv:2111.09543.

\bibitem[{Hendrycks et~al.(2021)Hendrycks, Burns, Basart, Zou, Mazeika, Song, and Steinhardt}]{MMLU}
Dan Hendrycks, Collin Burns, Steven Basart, Andy Zou, Mantas Mazeika, Dawn Song, and Jacob Steinhardt. 2021.
\newblock \href {http://dblp.uni-trier.de/db/conf/iclr/iclr2021.html#HendrycksBBZMSS21} {Measuring massive multitask language understanding.}
\newblock In \emph{ICLR}. OpenReview.net.

\bibitem[{Hernandez et~al.(2022)Hernandez, Brown, Conerly, DasSarma, Drain, El-Showk, Elhage, Hatfield-Dodds, Henighan, Hume, Johnston, Mann, Olah, Olsson, Amodei, Joseph, Kaplan, and McCandlish}]{RepeatedScaling}
Danny Hernandez, Tom Brown, Tom Conerly, Nova DasSarma, Dawn Drain, Sheer El-Showk, Nelson Elhage, Zac Hatfield-Dodds, Tom Henighan, Tristan Hume, Scott Johnston, Ben Mann, Chris Olah, Catherine Olsson, Dario Amodei, Nicholas Joseph, Jared Kaplan, and Sam McCandlish. 2022.
\newblock \href {https://arxiv.org/abs/2205.10487} {Scaling laws and interpretability of learning from repeated data}.
\newblock \emph{Preprint}, arXiv:2205.10487.

\bibitem[{Hoffmann et~al.(2022)Hoffmann, Borgeaud, Mensch, Buchatskaya, Cai, Rutherford, de~Las~Casas, Hendricks, Welbl, Clark, Hennigan, Noland, Millican, van~den Driessche, Damoc, Guy, Osindero, Simonyan, Elsen, Vinyals, Rae, and Sifre}]{Hoffmann}
Jordan Hoffmann, Sebastian Borgeaud, Arthur Mensch, Elena Buchatskaya, Trevor Cai, Eliza Rutherford, Diego de~Las~Casas, Lisa~Anne Hendricks, Johannes Welbl, Aidan Clark, Thomas Hennigan, Eric Noland, Katherine Millican, George van~den Driessche, Bogdan Damoc, Aurelia Guy, Simon Osindero, Kar\'{e}n Simonyan, Erich Elsen, Oriol Vinyals, Jack Rae, and Laurent Sifre. 2022.
\newblock \href {https://proceedings.neurips.cc/paper_files/paper/2022/file/c1e2faff6f588870935f114ebe04a3e5-Paper-Conference.pdf} {An empirical analysis of compute-optimal large language model training}.
\newblock In \emph{Advances in Neural Information Processing Systems}, volume~35, pages 30016--30030. Curran Associates, Inc.

\bibitem[{Jennings et~al.()Jennings, Patwary, Subramanian, Prabhumoye, Dattagupta, Jawa, Liu, Wolf, Yurick, and Singh}]{NeMo}
Joseph Jennings, Mostofa Patwary, Sandeep Subramanian, Shrimai Prabhumoye, Ayush Dattagupta, Vibhu Jawa, Jiwei Liu, Ryan Wolf, Sarah Yurick, and Varun Singh.
\newblock \href {https://github.com/NVIDIA/NeMo-Curator} {{NeMo-Curator: a toolkit for data curation}}.

\bibitem[{Joshi et~al.(2017)Joshi, Choi, Weld, and Zettlemoyer}]{triviaqa}
Mandar Joshi, Eunsol Choi, Daniel Weld, and Luke Zettlemoyer. 2017.
\newblock \href {https://doi.org/10.18653/v1/P17-1147} {{T}rivia{QA}: A large scale distantly supervised challenge dataset for reading comprehension}.
\newblock In \emph{Proceedings of the 55th Annual Meeting of the Association for Computational Linguistics (Volume 1: Long Papers)}, pages 1601--1611, Vancouver, Canada. Association for Computational Linguistics.

\bibitem[{Kang et~al.(2024)Kang, Sun, Wen, Chen, Song, Mahmood, and Jia}]{AutoScale}
Feiyang Kang, Yifan Sun, Bingbing Wen, Si~Chen, Dawn Song, Rafid Mahmood, and Ruoxi Jia. 2024.
\newblock \href {https://arxiv.org/abs/2407.20177} {Autoscale: Automatic prediction of compute-optimal data composition for training llms}.
\newblock \emph{Preprint}, arXiv:2407.20177.

\bibitem[{Kaplan et~al.(2020)Kaplan, McCandlish, Henighan, Brown, Chess, Child, Gray, Radford, Wu, and Amodei}]{ScalingLaws}
Jared Kaplan, Sam McCandlish, Tom Henighan, Tom~B. Brown, Benjamin Chess, Rewon Child, Scott Gray, Alec Radford, Jeffrey Wu, and Dario Amodei. 2020.
\newblock \href {https://arxiv.org/abs/2001.08361} {Scaling laws for neural language models}.
\newblock \emph{Preprint}, arXiv:2001.08361.

\bibitem[{Ke et~al.(2017)Ke, Meng, Finley, Wang, Chen, Ma, Ye, and Liu}]{lightgbm}
Guolin Ke, Qi~Meng, Thomas Finley, Taifeng Wang, Wei Chen, Weidong Ma, Qiwei Ye, and Tie-Yan Liu. 2017.
\newblock \href {https://proceedings.neurips.cc/paper_files/paper/2017/file/6449f44a102fde848669bdd9eb6b76fa-Paper.pdf} {Lightgbm: A highly efficient gradient boosting decision tree}.
\newblock In \emph{Advances in Neural Information Processing Systems}, volume~30. Curran Associates, Inc.

\bibitem[{Kwiatkowski et~al.(2019)Kwiatkowski, Palomaki, Redfield, Collins, Parikh, Alberti, Epstein, Polosukhin, Devlin, Lee, Toutanova, Jones, Kelcey, Chang, Dai, Uszkoreit, Le, and Petrov}]{NQ}
Tom Kwiatkowski, Jennimaria Palomaki, Olivia Redfield, Michael Collins, Ankur Parikh, Chris Alberti, Danielle Epstein, Illia Polosukhin, Jacob Devlin, Kenton Lee, Kristina Toutanova, Llion Jones, Matthew Kelcey, Ming-Wei Chang, Andrew~M. Dai, Jakob Uszkoreit, Quoc Le, and Slav Petrov. 2019.
\newblock \href {https://doi.org/10.1162/tacl_a_00276} {Natural questions: A benchmark for question answering research}.
\newblock \emph{Transactions of the Association for Computational Linguistics}, 7:452--466.

\bibitem[{Lauren\c{c}on et~al.(2022)Lauren\c{c}on, Saulnier, Wang, Akiki, Villanova~del Moral, Le~Scao, Von~Werra, Mou, Gonz\'{a}lez~Ponferrada, Nguyen, Frohberg, \v{S}a\v{s}ko, , and et~al}]{ROOT}
Hugo Lauren\c{c}on, Lucile Saulnier, Thomas Wang, Christopher Akiki, Albert Villanova~del Moral, Teven Le~Scao, Leandro Von~Werra, Chenghao Mou, Eduardo Gonz\'{a}lez~Ponferrada, Huu Nguyen, J\"{o}rg Frohberg, Mario \v{S}a\v{s}ko, , and et~al. 2022.
\newblock \href {https://proceedings.neurips.cc/paper_files/paper/2022/file/ce9e92e3de2372a4b93353eb7f3dc0bd-Paper-Datasets_and_Benchmarks.pdf} {The bigscience roots corpus: A 1.6tb composite multilingual dataset}.
\newblock In \emph{Advances in Neural Information Processing Systems}, volume~35, pages 31809--31826. Curran Associates, Inc.

\bibitem[{Lee et~al.(2019)Lee, Chang, and Toutanova}]{NQ_open}
Kenton Lee, Ming-Wei Chang, and Kristina Toutanova. 2019.
\newblock \href {https://doi.org/10.18653/v1/P19-1612} {Latent retrieval for weakly supervised open domain question answering}.
\newblock In \emph{Proceedings of the 57th Annual Meeting of the Association for Computational Linguistics}, pages 6086--6096, Florence, Italy. Association for Computational Linguistics.

\bibitem[{Li et~al.(2024)Li, Fang, Smyrnis, Ivgi, Jordan, Gadre, Bansal, Guha, Keh, and et~al}]{dclm}
Jeffrey Li, Alex Fang, Georgios Smyrnis, Maor Ivgi, Matt Jordan, Samir Gadre, Hritik Bansal, Etash Guha, Sedrick Keh, and et~al. 2024.
\newblock \href {https://arxiv.org/abs/2406.11794} {Datacomp-lm: In search of the next generation of training sets for language models}.
\newblock \emph{Preprint}, arXiv:2406.11794.

\bibitem[{Lin et~al.(2025)Lin, Gou, Gong, Liu, Shen, Xu, Lin, Yang, Jiao, Duan, and Chen}]{Rho}
Zhenghao Lin, Zhibin Gou, Yeyun Gong, Xiao Liu, Yelong Shen, Ruochen Xu, Chen Lin, Yujiu Yang, Jian Jiao, Nan Duan, and Weizhu Chen. 2025.
\newblock \href {https://arxiv.org/abs/2404.07965} {Rho-1: Not all tokens are what you need}.
\newblock \emph{Preprint}, arXiv:2404.07965.

\bibitem[{Liu et~al.(2024)Liu, Zheng, Muennighoff, Zeng, Dou, Pang, Jiang, and Lin}]{regmix}
Qian Liu, Xiaosen Zheng, Niklas Muennighoff, Guangtao Zeng, Longxu Dou, Tianyu Pang, Jing Jiang, and Min Lin. 2024.
\newblock \href {https://arxiv.org/abs/2407.01492} {Regmix: Data mixture as regression for language model pre-training}.
\newblock \emph{Preprint}, arXiv:2407.01492.

\bibitem[{Longpre et~al.(2024)Longpre, Yauney, Reif, Lee, Roberts, Zoph, Zhou, Wei, Robinson, Mimno, and Ippolito}]{Longpre:2024}
Shayne Longpre, Gregory Yauney, Emily Reif, Katherine Lee, Adam Roberts, Barret Zoph, Denny Zhou, Jason Wei, Kevin Robinson, David Mimno, and Daphne Ippolito. 2024.
\newblock \href {https://doi.org/10.18653/v1/2024.naacl-long.179} {A pretrainer{'}s guide to training data: Measuring the effects of data age, domain coverage, quality, {\&} toxicity}.
\newblock In \emph{Proceedings of the 2024 Conference of the North American Chapter of the Association for Computational Linguistics: Human Language Technologies (Volume 1: Long Papers)}, pages 3245--3276, Mexico City, Mexico. Association for Computational Linguistics.

\bibitem[{Marion et~al.(2023)Marion, Üstün, Pozzobon, Wang, Fadaee, and Hooker}]{LIMO}
Max Marion, Ahmet Üstün, Luiza Pozzobon, Alex Wang, Marzieh Fadaee, and Sara Hooker. 2023.
\newblock \href {https://arxiv.org/abs/2309.04564} {When less is more: Investigating data pruning for pretraining llms at scale}.
\newblock \emph{Preprint}, arXiv:2309.04564.

\bibitem[{Mihaylov et~al.(2018)Mihaylov, Clark, Khot, and Sabharwal}]{OpenBookQA}
Todor Mihaylov, Peter Clark, Tushar Khot, and Ashish Sabharwal. 2018.
\newblock \href {https://doi.org/10.18653/v1/D18-1260} {Can a suit of armor conduct electricity? a new dataset for open book question answering}.
\newblock In \emph{Proceedings of the 2018 Conference on Empirical Methods in Natural Language Processing}, pages 2381--2391, Brussels, Belgium. Association for Computational Linguistics.

\bibitem[{Muennighoff et~al.(2023)Muennighoff, Rush, Barak, Le~Scao, Tazi, Piktus, Pyysalo, Wolf, and Raffel}]{Muennighoff}
Niklas Muennighoff, Alexander Rush, Boaz Barak, Teven Le~Scao, Nouamane Tazi, Aleksandra Piktus, Sampo Pyysalo, Thomas Wolf, and Colin~A Raffel. 2023.
\newblock \href {https://proceedings.neurips.cc/paper_files/paper/2023/file/9d89448b63ce1e2e8dc7af72c984c196-Paper-Conference.pdf} {Scaling data-constrained language models}.
\newblock In \emph{Advances in Neural Information Processing Systems}, volume~36, pages 50358--50376. Curran Associates, Inc.

\bibitem[{Patel et~al.(2021)Patel, Bhattamishra, and Goyal}]{SVAMP}
Arkil Patel, Satwik Bhattamishra, and Navin Goyal. 2021.
\newblock \href {https://doi.org/10.18653/v1/2021.naacl-main.168} {Are {NLP} models really able to solve simple math word problems?}
\newblock In \emph{Proceedings of the 2021 Conference of the North American Chapter of the Association for Computational Linguistics: Human Language Technologies}, pages 2080--2094, Online. Association for Computational Linguistics.

\bibitem[{Penedo et~al.(2024)Penedo, Kydlíček, allal, Lozhkov, Mitchell, Raffel, Werra, and Wolf}]{fineweb-edu}
Guilherme Penedo, Hynek Kydlíček, Loubna~Ben allal, Anton Lozhkov, Margaret Mitchell, Colin Raffel, Leandro~Von Werra, and Thomas Wolf. 2024.
\newblock \href {https://arxiv.org/abs/2406.17557} {The fineweb datasets: Decanting the web for the finest text data at scale}.
\newblock \emph{Preprint}, arXiv:2406.17557.

\bibitem[{Penedo et~al.(2023)Penedo, Malartic, Hesslow, Cojocaru, Alobeidli, Cappelli, Pannier, Almazrouei, and Launay}]{refinedweb}
Guilherme Penedo, Quentin Malartic, Daniel Hesslow, Ruxandra Cojocaru, Hamza Alobeidli, Alessandro Cappelli, Baptiste Pannier, Ebtesam Almazrouei, and Julien Launay. 2023.
\newblock \href {https://proceedings.neurips.cc/paper_files/paper/2023/file/fa3ed726cc5073b9c31e3e49a807789c-Paper-Datasets_and_Benchmarks.pdf} {The refinedweb dataset for falcon llm: Outperforming curated corpora with web data only}.
\newblock In \emph{Advances in Neural Information Processing Systems}, volume~36, pages 79155--79172. Curran Associates, Inc.

\bibitem[{Que et~al.(2024)Que, Liu, Zhang, Zhang, Qu, Ma, Duan, Bai, Wang, Zhang, Tan, Fu, Su, Wang, Qu, and Zheng}]{D-CPT}
Haoran Que, Jiaheng Liu, Ge~Zhang, Chenchen Zhang, Xingwei Qu, Yinghao Ma, Feiyu Duan, Zhiqi Bai, Jiakai Wang, Yuanxing Zhang, Xu~Tan, Jie Fu, Wenbo Su, Jiamang Wang, Lin Qu, and Bo~Zheng. 2024.
\newblock \href {https://arxiv.org/abs/2406.01375} {D-cpt law: Domain-specific continual pre-training scaling law for large language models}.
\newblock \emph{Preprint}, arXiv:2406.01375.

\bibitem[{Rae et~al.(2022)Rae, Borgeaud, Cai, Millican, Hoffmann, Song, Aslanides, Henderson, Ring, Young, Rutherford, Hennigan, Menick, Cassirer, Powell, van~den Driessche, Hendricks, Rauh, Huang, Glaese, Welbl, Dathathri, and et~al}]{ScalingLanguageModels}
Jack~W. Rae, Sebastian Borgeaud, Trevor Cai, Katie Millican, Jordan Hoffmann, Francis Song, John Aslanides, Sarah Henderson, Roman Ring, Susannah Young, Eliza Rutherford, Tom Hennigan, Jacob Menick, Albin Cassirer, Richard Powell, George van~den Driessche, Lisa~Anne Hendricks, Maribeth Rauh, Po-Sen Huang, Amelia Glaese, Johannes Welbl, Sumanth Dathathri, and et~al. 2022.
\newblock \href {https://arxiv.org/abs/2112.11446} {Scaling language models: Methods, analysis {\&} insights from training gopher}.
\newblock \emph{Preprint}, arXiv:2112.11446.

\bibitem[{Raffel et~al.(2020)Raffel, Shazeer, Roberts, Lee, Narang, Matena, Zhou, Li, and Liu}]{C4}
Colin Raffel, Noam Shazeer, Adam Roberts, Katherine Lee, Sharan Narang, Michael Matena, Yanqi Zhou, Wei Li, and Peter~J. Liu. 2020.
\newblock \href {http://jmlr.org/papers/v21/20-074.html} {Exploring the limits of transfer learning with a unified text-to-text transformer}.
\newblock \emph{Journal of Machine Learning Research}, 21(140):1--67.

\bibitem[{Sachdeva et~al.(2024)Sachdeva, Coleman, Kang, Ni, Hong, Chi, Caverlee, McAuley, and Cheng}]{askllm}
Noveen Sachdeva, Benjamin Coleman, Wang-Cheng Kang, Jianmo Ni, Lichan Hong, Ed~H. Chi, James Caverlee, Julian McAuley, and Derek~Zhiyuan Cheng. 2024.
\newblock \href {https://arxiv.org/abs/2402.09668} {How to train data-efficient llms}.
\newblock \emph{Preprint}, arXiv:2402.09668.

\bibitem[{Shao et~al.(2024)Shao, Li, Fei, Yan, Lin, and Qiu}]{Clusterclip}
Yunfan Shao, Linyang Li, Zhaoye Fei, Hang Yan, Dahua Lin, and Xipeng Qiu. 2024.
\newblock \href {https://arxiv.org/abs/2402.14526} {Balanced data sampling for language model training with clustering}.
\newblock \emph{Preprint}, arXiv:2402.14526.

\bibitem[{Shazeer(2020)}]{SwiGLU}
Noam Shazeer. 2020.
\newblock \href {https://arxiv.org/abs/2002.05202} {Glu variants improve transformer}.
\newblock \emph{Preprint}, arXiv:2002.05202.

\bibitem[{Soldaini et~al.(2024)Soldaini, Kinney, Bhagia, Schwenk, Atkinson, Authur, Bogin, Chandu, Dumas, Elazar, Hofmann, Jha, Kumar, Lucy, Lyu, Lambert, and et~al}]{Dolma}
Luca Soldaini, Rodney Kinney, Akshita Bhagia, Dustin Schwenk, David Atkinson, Russell Authur, Ben Bogin, Khyathi Chandu, Jennifer Dumas, Yanai Elazar, Valentin Hofmann, Ananya Jha, Sachin Kumar, Li~Lucy, Xinxi Lyu, Nathan Lambert, and et~al. 2024.
\newblock \href {https://doi.org/10.18653/v1/2024.acl-long.840} {Dolma: an open corpus of three trillion tokens for language model pretraining research}.
\newblock In \emph{Proceedings of the 62nd Annual Meeting of the Association for Computational Linguistics (Volume 1: Long Papers)}, pages 15725--15788, Bangkok, Thailand. Association for Computational Linguistics.

\bibitem[{Su et~al.(2024)Su, Ahmed, Lu, Pan, Bo, and Liu}]{ROPE}
Jianlin Su, Murtadha Ahmed, Yu~Lu, Shengfeng Pan, Wen Bo, and Yunfeng Liu. 2024.
\newblock \href {https://doi.org/10.1016/j.neucom.2023.127063} {Roformer: Enhanced transformer with rotary position embedding}.
\newblock \emph{Neurocomputing}, 568:127063.

\bibitem[{Teknium(2023)}]{Openhermes}
Teknium. 2023.
\newblock \href {https://huggingface.co/datasets/teknium/OpenHermes-2.5} {Openhermes 2.5: An open dataset of synthetic data for generalist llm assistants}.
\newblock In \emph{huggingface}.

\bibitem[{Thrush et~al.(2024)Thrush, Potts, and Hashimoto}]{Tristan:2024}
Tristan Thrush, Christopher Potts, and Tatsunori Hashimoto. 2024.
\newblock \href {https://arxiv.org/abs/2409.05816} {Improving pretraining data using perplexity correlations}.
\newblock \emph{Preprint}, arXiv:2409.05816.

\bibitem[{Tirumala et~al.(2023)Tirumala, Simig, Aghajanyan, and Morcos}]{D4}
Kushal Tirumala, Daniel Simig, Armen Aghajanyan, and Ari~S. Morcos. 2023.
\newblock \href {https://arxiv.org/abs/2308.12284} {D4: Improving llm pretraining via document de-duplication and diversification}.
\newblock \emph{Preprint}, arXiv:2308.12284.

\bibitem[{Touvron et~al.(2023)Touvron, Lavril, Izacard, Martinet, Lachaux, Lacroix, Rozière, Goyal, Hambro, Azhar, Rodriguez, Joulin, Grave, and Lample}]{llama}
Hugo Touvron, Thibaut Lavril, Gautier Izacard, Xavier Martinet, Marie-Anne Lachaux, Timothée Lacroix, Baptiste Rozière, Naman Goyal, Eric Hambro, Faisal Azhar, Aurelien Rodriguez, Armand Joulin, Edouard Grave, and Guillaume Lample. 2023.
\newblock \href {https://arxiv.org/abs/2302.13971} {Llama: Open and efficient foundation language models}.
\newblock \emph{Preprint}, arXiv:2302.13971.

\bibitem[{Vaswani et~al.(2017)Vaswani, Shazeer, Parmar, Uszkoreit, Jones, Gomez, Kaiser, and Polosukhin}]{Attention}
Ashish Vaswani, Noam Shazeer, Niki Parmar, Jakob Uszkoreit, Llion Jones, Aidan~N Gomez, \L~ukasz Kaiser, and Illia Polosukhin. 2017.
\newblock \href {https://proceedings.neurips.cc/paper_files/paper/2017/file/3f5ee243547dee91fbd053c1c4a845aa-Paper.pdf} {Attention is all you need}.
\newblock In \emph{Advances in Neural Information Processing Systems}, volume~30. Curran Associates, Inc.

\bibitem[{Weber et~al.(2024)Weber, Fu, Anthony, Oren, Adams, Alexandrov, Lyu, Nguyen, Yao, Adams, Athiwaratkun, Chalamala, Chen, Ryabinin, Dao, Liang, Ré, Rish, and Zhang}]{RedPajama}
Maurice Weber, Daniel Fu, Quentin Anthony, Yonatan Oren, Shane Adams, Anton Alexandrov, Xiaozhong Lyu, Huu Nguyen, Xiaozhe Yao, Virginia Adams, Ben Athiwaratkun, Rahul Chalamala, Kezhen Chen, Max Ryabinin, Tri Dao, Percy Liang, Christopher Ré, Irina Rish, and Ce~Zhang. 2024.
\newblock \href {https://arxiv.org/abs/2411.12372} {Redpajama: an open dataset for training large language models}.
\newblock \emph{Preprint}, arXiv:2411.12372.

\bibitem[{Wenzek et~al.(2020)Wenzek, Lachaux, Conneau, Chaudhary, Guzm\'an, Joulin, and Grave}]{ccnet}
Guillaume Wenzek, Marie-Anne Lachaux, Alexis Conneau, Vishrav Chaudhary, Francisco Guzm\'an, Armand Joulin, and Edouard Grave. 2020.
\newblock \href {https://aclanthology.org/2020.lrec-1.494/} {{CCN}et: Extracting high quality monolingual datasets from web crawl data}.
\newblock In \emph{Proceedings of the Twelfth Language Resources and Evaluation Conference}, pages 4003--4012, Marseille, France. European Language Resources Association.

\bibitem[{Wettig et~al.(2024)Wettig, Gupta, Malik, and Chen}]{QuRating}
Alexander Wettig, Aatmik Gupta, Saumya Malik, and Danqi Chen. 2024.
\newblock \href {https://arxiv.org/abs/2402.09739} {Qurating: Selecting high-quality data for training language models}.
\newblock \emph{Preprint}, arXiv:2402.09739.

\bibitem[{Xie et~al.(2023{\natexlab{a}})Xie, Pham, Dong, Du, Liu, Lu, Liang, Le, Ma, and Yu}]{doremi}
Sang~Michael Xie, Hieu Pham, Xuanyi Dong, Nan Du, Hanxiao Liu, Yifeng Lu, Percy~S Liang, Quoc~V Le, Tengyu Ma, and Adams~Wei Yu. 2023{\natexlab{a}}.
\newblock \href {https://proceedings.neurips.cc/paper_files/paper/2023/file/dcba6be91359358c2355cd920da3fcbd-Paper-Conference.pdf} {Doremi: Optimizing data mixtures speeds up language model pretraining}.
\newblock In \emph{Advances in Neural Information Processing Systems}, volume~36, pages 69798--69818. Curran Associates, Inc.

\bibitem[{Xie et~al.(2023{\natexlab{b}})Xie, Santurkar, Ma, and Liang}]{DSIR}
Sang~Michael Xie, Shibani Santurkar, Tengyu Ma, and Percy~S Liang. 2023{\natexlab{b}}.
\newblock \href {https://proceedings.neurips.cc/paper_files/paper/2023/file/6b9aa8f418bde2840d5f4ab7a02f663b-Paper-Conference.pdf} {Data selection for language models via importance resampling}.
\newblock In \emph{Advances in Neural Information Processing Systems}, volume~36, pages 34201--34227. Curran Associates, Inc.

\bibitem[{Ye et~al.(2024)Ye, Liu, Sun, Zhou, Zhan, and Qiu}]{DML}
Jiasheng Ye, Peiju Liu, Tianxiang Sun, Yunhua Zhou, Jun Zhan, and Xipeng Qiu. 2024.
\newblock \href {https://arxiv.org/abs/2403.16952} {Data mixing laws: Optimizing data mixtures by predicting language modeling performance}.
\newblock \emph{Preprint}, arXiv:2403.16952.

\bibitem[{Yu et~al.(2024)Yu, Das, and Xiong}]{MATES}
Zichun Yu, Spandan Das, and Chenyan Xiong. 2024.
\newblock \href {https://arxiv.org/abs/2406.06046} {Mates: Model-aware data selection for efficient pretraining with data influence models}.
\newblock \emph{Preprint}, arXiv:2406.06046.

\bibitem[{Zellers et~al.(2019)Zellers, Holtzman, Bisk, Farhadi, and Choi}]{hellaswag}
Rowan Zellers, Ari Holtzman, Yonatan Bisk, Ali Farhadi, and Yejin Choi. 2019.
\newblock \href {https://doi.org/10.18653/v1/P19-1472} {{H}ella{S}wag: Can a machine really finish your sentence?}
\newblock In \emph{Proceedings of the 57th Annual Meeting of the Association for Computational Linguistics}, pages 4791--4800, Florence, Italy. Association for Computational Linguistics.

\end{thebibliography}

\end{document}